\documentclass[review,authoryear,10pt]{elsarticle}

\usepackage{latexsym}
\usepackage{amssymb}
\usepackage[T1]{fontenc}
\usepackage{amsmath}
\usepackage{graphicx}
\usepackage{amsmath}
\usepackage{xcolor,float}
\usepackage{soul}
\usepackage{multirow}
\usepackage[left=1in,top=1in,right=1in,bottom=1in,nohead]{geometry}
\usepackage{lineno}

\journal{ArXiv}

\begin{document}
	
	\begin{frontmatter}
	\title{High-Throughput Image-Based Plant Stand Count Estimation Using Convolutional Neural Networks}

	\begin{abstract}
    
   The future landscape of modern farming and plant breeding is rapidly changing due to the complex needs of our society. The explosion of collectable data has started a revolution in agriculture to the point where innovation must occur. To a commercial organization, the accurate and efficient collection of information is necessary to ensure that optimal decisions are made at key points of the breeding cycle. However, due to the shear size of a breeding program and current resource limitations, the ability to collect precise data on individual plants is not possible. In particular, efficient phenotyping of crops to record its color, shape, chemical properties, disease susceptibility, etc. is severely limited due to labor requirements and, oftentimes, expert domain knowledge. In this paper, we propose a deep learning based approach, named DeepStand, for image-based corn stand counting at early phenological stages. The proposed method adopts a truncated VGG-16 network as a backbone feature extractor and merges multiple feature maps with different scales to make the network robust against scale variation. Our extensive computational experiments suggest that our proposed method can successfully count corn stands and out-perform other state-of-the-art methods. It is the goal of our work to be used by the larger agricultural community as a way to enable high-throughput phenotyping without the use of extensive time and labor requirements.

	\end{abstract}
	
	\begin{keyword}
	Corn stand counting, Convolutional neural networks,  Plant phenotyping, Deep learning
	\end{keyword}

	\author[ISU]{Saeed Khaki\corref{cor1}}
	\ead{skhaki@iastate.edu}	
	
	\author[SYN]{Hieu Pham}
	
	\author[SYN]{Ye Han}
	
	\author[SYN]{Wade Kent}
	
	\author[ISU]{Lizhi Wang}
	
	
	\address[ISU]{Department of Industrial and Manufacturing Systems Engineering, Iowa State University, Ames, Iowa USA}
	
	\address[SYN]{Syngenta Seeds, Slater, Iowa USA}	
	
	\end{frontmatter}

\section{Introduction}

The ``phenotyping bottleneck'' refers to the phenomenon that the agricultural community is not able to accurately and efficiently collect data on physical properties of crops \citep{furbank2011phenomics}. This high-throughput phenotyping (HTP) bottleneck is oftentimes due to resource limitations such as labor, time, and domain expertise in pathology and genetics, even for large scale farming operations and commercial breeding programs. Certain phenotypic traits such as early stage stand counting for corn (Zea $mays$ L.) can only be accurately performed during its early growth stage. If this data collection time-window is missed, then the task is near impossible to complete \citep{mcwilliams1999corn}.

Understanding the data collection challenges facing modern agriculture, agronomists and researchers have turned to modern solutions that combine modern machine learning and imaging analytics \citep{yang2020crop}. Image-capturing devices such as unmanned aerial vehicles, high definition cameras, and, even, cell phone cameras are being used as devices to collect information to be analyzed either live or at a later time \citep{mogili2018review, kulbacki2018survey}. With these new technologies organizations and farmers are no longer bounded by the data collection time-window and now have the ability to manually analyze images at a later time. However, with this new approach comes additional problems such as storing massive amounts of image-based data and a familiar but new challenge - analyzing massive reserves of images accurately and efficiently. To advance modern agriculture, tools must be made easily available to agronomists to enable real-time decision making. The information contained within these images allows for timely management decision to optimize yield against harmful attack vectors (pests, diseases, drought, etc.).

To analyze large reserves of images quickly modern deep learning tools have been invoked by various crops. Recent literature has seen the combination of planting phenotyping and traditional machine learning techniques to count crops, detect color and classify stress in various crops through images \citep{singh2016machine, naik2017real, yuan2018wheat, guo2018aerial}. These recent works demonstrate the impact traditional machine learning has on the future of agriculture. However, these methods are not without limitations. Using traditional approaches oftentimes, requires high quality images, constant lighting conditions, and fixed camera distances. These limitations act as a barrier to true HTP. With advances in state-of-the-art deep learning techniques, these constraints are no longer binding. Robust models can be constructed to analyze crops in numerous variable conditions. This is seen in the current literature combining deep learning and HTP.

Image-based phenotyping and deep learning can broadly be labeled as an application area of computer vision. Traditional tasks include classifying single images, counting objects and detection objects. Common deep learning frameworks using AlexNet, LeNet, and ResNet-50 architectures have been applied to classify various fruits and vegetables from single images \citep{mohanty2016using, cruz2017x,wang2017automatic}. Other deep learning models using VGG-16 as a feature extractor and the ``You Only Look Once'' model has been used to count and detect leaves, sorghum heads, and corn kernels \citep{giuffrida2018pheno, ghosal2019weakly,mosley2020image, khaki2020deepcorn, redmon2016you}. Using novel frameworks to count corn tassels, \cite{lu2017tasselnet} combined convolutional neural networks (CNN) and local counts regression into a framework they call TasselNet. Additionally, open-access, high-quality, annotated datasets are being created and released to the public to engage researchers in applying their deep learning knowledge to agriculture \citep{zheng2019cropdeep,sudars2020dataset,haug2014crop}. These recent works showcase the potential for combining modern deep learning and agriculture in hopes of mitigating the so called ``phenotyping bottleneck''. For the curious reader who is interested in a clear, concise, and thorough review of image-based HTP, we point the reader towards a survey paper by \cite{jiang2020convolutional}.

Corn is known to be one of the world's most essential crop due to the number of products that it can create (e.g. flour, bio-fuels) \citep{berardi2019flooding}. Additionally, a large percentage of corn is used in livestock farming to feed pigs, cattle, and cows. The world's reliance on corn cannot be understated. Aside from the manufacturing aspect, corn has a large impact on the United States' economy. In 2019, it is estimated that the U.S. corn market contributed approximately \$140 billion to the U.S. economy. It is evident that the agricultural community and the world must act to ensure the continued optimal production of corn. By 2050, the world's population is estimated to be approximately 9 billion \citep{stephenson2010population}. The increase in population combined with the non-increasing arable land, changes will need to occur so that we can continue to optimize corn yield while utilizing less resources. Previous studies invoke deep learning based approaches to  predict corn yield based using genetics, environment, and satellite imagery, but these studies are not considered HTP on commercial corn and only act as a way to estimate yield during the growing season \citep{khaki2019classification,russello2018convolutional,khaki2019cnn,khaki2019crop,khaki2020predicting}.

The ultimate goal of this paper is to count the number of corn stands in an image of a specific area on the field taken during the early phenological stages (VE to V6) \citep{ciampitti2011corn}. Roughly these phenological stages refer to the visible leaves on the stem. For instance, VE (emergence) is the first phenological stage where the stem breaks through the soil. V1 is the appearance of the first leaf. V2 is appearance of the second leaf, and VN is the appearance of the $n$-th leaf. From a practical perspective, an estimated stand count value allows for the establishment of a planting rate and, ultimately, yield potential. If the proper rate is planted, then farmers/breeders can estimate yield based on product by population. However, if population is not there i.e. poor germination/ bad planter, then farmers can identify the issue and replant, or at the very least establish what the new yield will be. Knowing that the planting density is below its desired threshold enables farmers to decide how they want to best manage their corn to make up for the difference in planting rating (e.g. more fertilizer, more aggressive pesticide control, etc.).  Traditionally, farmers perform stand count estimations manually. However, this process is time consuming, labor intensive and prone to human error. Because of this, there is a reluctance to perform a stand count, ultimately, leaving farmers at a net-loss for the corn yield. Utilizing an image-based approach to this problem will allow for the timely estimation of stand counts and well as a consistent measure to the quality of data.

Due to the need of efficient and effective HTP, in this paper, we present a deep learning based approach, named DeepStand, to alleviate the concerns of manual, labor intensive stand counting. The proposed method adopts a truncated VGG-16 network as a backbone feature extractor and merges multiple feature maps with different scales to make the network robust against scale variations. This approach is similar to common methods in crowd counting where models are used to detect individual people in large crowds. Due to the similarity of these problems, we utilize a point density based approach for detecting the corn stands.

\section{Methodology}\label{sec:method}

\begin{figure}[H]
    \centering
    \includegraphics[scale=0.25]{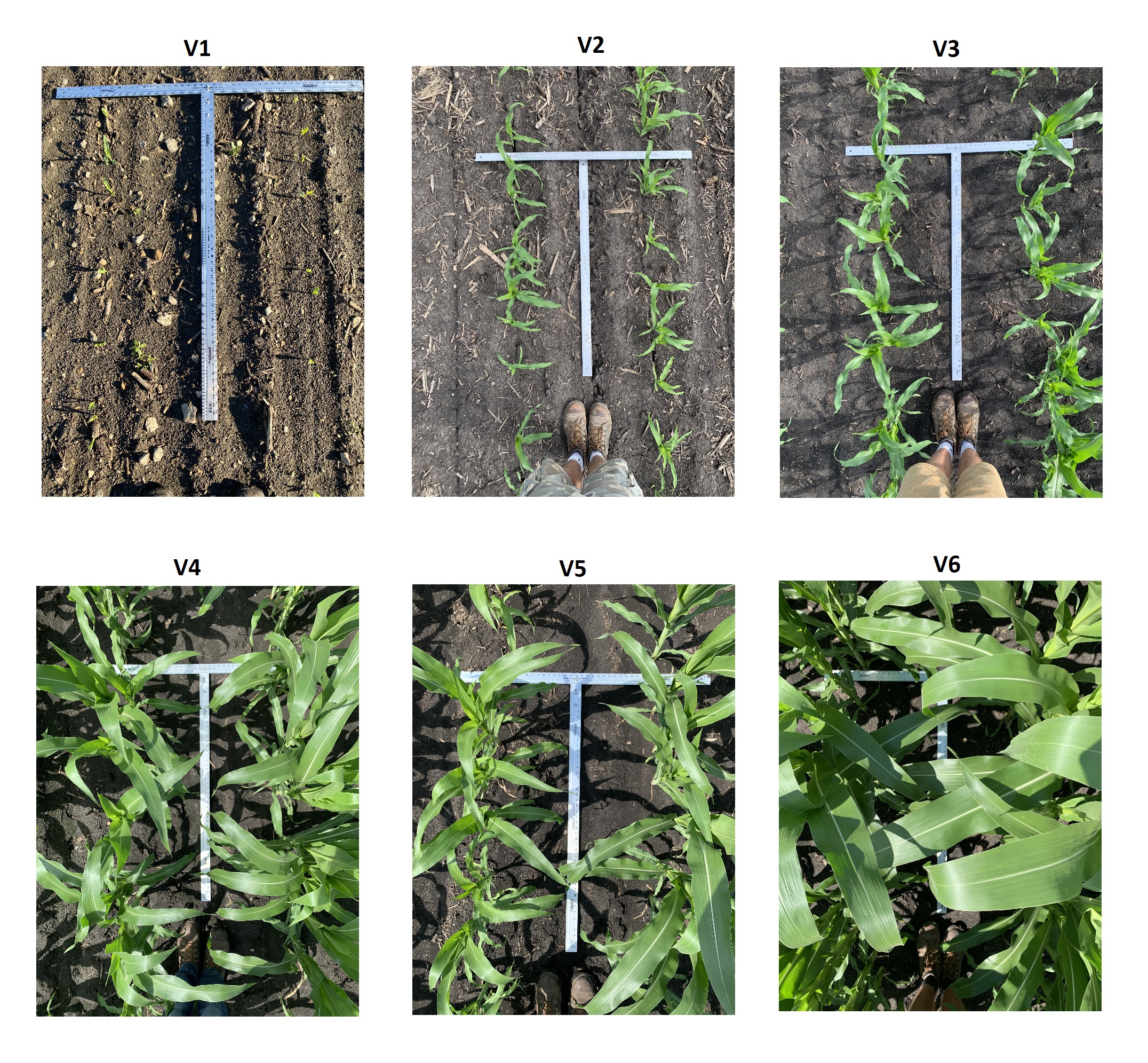}
    \caption{The images show corn stands at vegetative stages V1 to V6. The images include high scale variations, occlusions, and small distance between corn stands especially at stages V4 to V6.}
    \label{fig:corn_stand}
\end{figure}

Image-based corn stand counting is challenging compared to the counting tasks in other fields due to multiple factors, including occlusions, scale variations, and small distance between corn stands. Figure \ref{fig:corn_stand} shows the corn stands at different growing stages. This paper proposes a deep learning based method, DeepStand, to count the number of corn stands based on using a 180-degree image taken at 4-6 feet above the ground. It is worth mentioning that as corn progresses through its phenological stages, accurately counting the planting density becomes a difficult task for a computer due to the amount of overlapping leaves. However, from a pragmatic perspective, stand counting should be performed before V4 to ensure agronomists can act in a timely manner to mitigate any crop issues.

\subsection{Network Architecture}

Corn stand images usually include high scale variations and occluded corn stands by nearby corn leaves. As a result, the proposed counting method should be robust against these factors. Our proposed stand counting method is inspired by methods proposed for the counting task in other fields such as crowd counting \citep{gao2020cnn} and dense object counting \citep{gao2020counting}. 

\begin{figure}[H]
    \centering
    \includegraphics[scale=0.2]{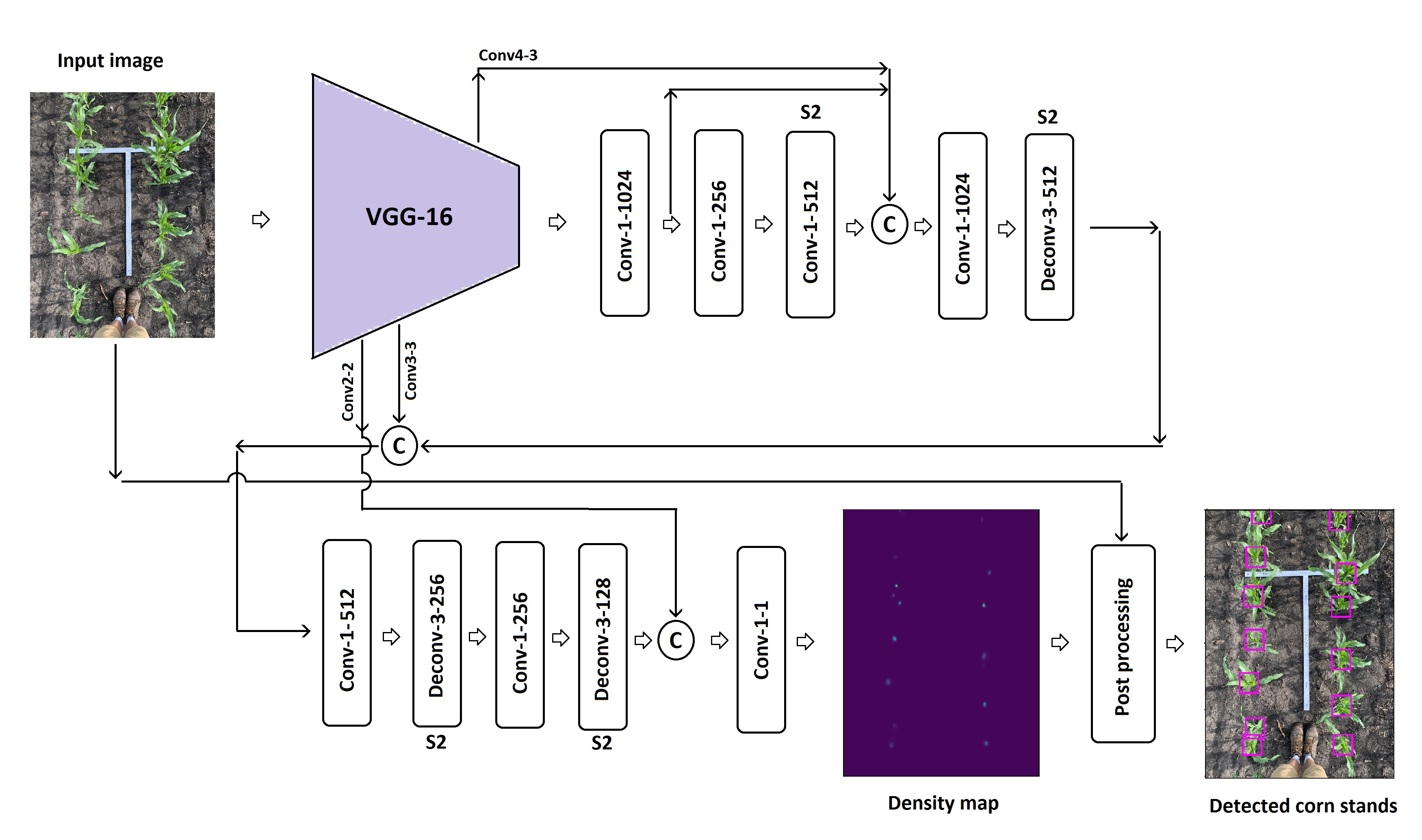}
    \caption{The outline of the DeepStand architecture. The parameters of the convolutional layers and deconvolutional layers are denoted as 'Conv-(kernel size)-(number of filters)' and 'Deconv-(kernel size)-(number of filters)', respectively. All layers have the stride of 1 except for the layers with "S2" notation, which have stride of 2. The padding type is the 'same' except for the first deconvolutional layer for which we use 'valid' padding. \textcircled{\raisebox{-0.8pt}{c}} denotes matrix concatenation.}
    \label{fig:model}
\end{figure}

The architecture of the proposed network is outlined in Figure \ref{fig:model}. The proposed network generates a density map given an image of corn stands, where integral over the density map gives the total number of corn stands. Our proposed method is a CNN-based density estimation method. We do not use other methods such as detection-based \citep{li2008estimating,zeng2010robust} or regression-based \citep{chan2008privacy,idrees2013multi,wang2015deep} methods for the following main reasons. Detection-based approaches usually apply an object detection method such as faster R-CNN \citep{ren2015faster}, SSD \citep{liu2016ssd} or a detector via a sliding window \citep{khaki2020convolutional} on an image to first detect the objects and then count them. However, theses approaches may not work well when applied to the scenes with occlusion and dense objects. Moreover, training these methods requires  a considerable amount of annotated images, which is not publicly available for the task of corn stand counting. Regression-based approaches directly map an image patch to the count. These approaches deal with the problems of occlusion and background clutter successfully, however, they ignore the spatial information in the input image. As such, these approaches do not know how much each region of image contribute the final count \citep{gao2020counting}.

We use a truncated VGG-16 \citep{simonyan2014deep} as a backbone for feature extraction in our proposed network. The truncated VGG-16 is composed of convolutional layers with a fixed kernel size of $3\times3$ that extracts discriminative features from input image for further analysis of the network. We use the VGG-16 network backbone in our network mainly because of its good generalization ability to other vision tasks such as counting and object detection \citep{liu2016ssd,gao2020counting,liu2019recurrent,li2018csrnet}. The truncated VGG-16 network includes all layers of VGG-16 network except the last max-pooling layer and all fully connected layer. The truncated VGG-16 shrinks the input images' resolution to the $1/8$ of its original size. We increase input size of the truncated VGG-16 network form $224\times224$ to $300\times300$ in our proposed network to learn more fine-grained features and patterns from the input image to further improve accuracy of our proposed method \citep{tan2019efficientnet}.

The proposed network merges feature maps from multiple scales of the network to make it robust against scale variations in images. Similar scale-adaptive architectures have been used in other vision studies \citep{zhang2018crowd,ronneberger2015u,bai2019crowd}. To concatenate feature maps with different spatial resolutions, we use zero padding to enlarge the smaller feature maps to the size of the largest feature map. Finally, we use three deconvolutional layers (transposed convolution) \citep{dumoulin2016guide} with stride of 2 to up-sample the output of the network to the size of the original input image. 

Finally, we do a post processing on the predicted density map to draw a bounding box around each corn stand. The post processing includes the following steps: (1) threshold the estimated density map to zero-out regions where the value of density map is insignificant, (2) find peak coordinates on the density map as the center location of corn stands, (3) draw a bounding box around each corn stand, and (4) apply non-maximum suppression to remove overlapping bounding boxes. The above-mentioned post processing has a very low computational cost and does not increase the inference time.

\subsection{Loss Function}

Let $I_i$, $D_i$, $F(I_i,\Theta)$, $\Theta$,  and $N$ denote $i$th image, the $i$th ground truth density map, the predicted density map of the $i$th image, network parameters, and the number of input images, respectively. As such, the network loss can be defined as below:

\begin{eqnarray}
L(\Theta)=\frac{1}{N}\sum_{i=1}^{N}\|F(I_i,\Theta)-D_i\|_{2}^{2}\label{eq:eq_loss}
\end{eqnarray}

Euclidean loss measures estimation error at pixel level and has been used in other crowd counting studies \citep{boominathan2016crowdnet,gao2020counting,lian2019density}.

\section{Experiments and Results}\label{sec:experiments}

This section first introduces the dataset used in our study, data augmentation, evaluation metrics, and training procedure. Then, we report the results of our proposed method along with other competing methods. All experiments were conducted in Tensorflow framework \citep{abadi2016tensorflow} on a NVIDIA Tesla V100 SXM2 GPU.

\subsection{Data}

\subsubsection{Ground Truth Density Maps Generation}

We generate ground truth density maps following the procedure of density map generation in \cite{boominathan2016crowdnet} to train the network parameters. A corn stand located at pixel $x_i$ can be represented by a delta function $\delta(x-x_i)$. As such, ground truth output for an image with $M$ annotated corn stands can be defined as follows:

\begin{eqnarray}
H(x)=\sum_{i=1}^{M}\delta(x-x_i)\label{eq:densitymap1}
\end{eqnarray}

\noindent Then, the $H(x)$ is convoluted with a Guassian kernel with a standard deviation $\sigma$ to generate the density map $D(x)$, where the standard deviation can be defined based on the average distance of k-nearest neighboring annotations. The summation over the density map is equal to the number of corn stands presented in the image.

\begin{eqnarray}
D(x)=\sum_{i=1}^{M}\delta(x-x_i) \ast G_\sigma(x)\label{eq:eq_2}
\end{eqnarray}

\noindent The use of such density maps as ground truth can help CNNs learn from the spatial information in images.

\subsubsection{Stand Count Data}\label{sec:data_aug}

This section presents the procedure to prepare sufficient data to train and evaluate our proposed method. Our original dataset includes 394 images of corn stands at growing stages V1 to V6 with a fixed size of $1024\times768$ taken at 4-6 feet above the ground. This includes a total of 6154 total stands across all images. The following table shows the summary of statistics of the dataset.

\begin{table}[H]
    \centering
    \begin{tabular}{|c|c|c|c|c|c|}

    \hline
       Number of Images  & Resolution & Min & Max& Avg & Total \\
       \hline
    \hline
      394   & $1024\times768$ & 5& 31 & 15.62& 6154\\
      \hline
    \end{tabular}
    \caption{The summary statistics of the corn stand dataset. The Min, Max, Avg, and Total denote the the minimum, maximum, average, and total number of corn stands in the dataset, respectively  }
    \label{tab:summary_dataset}
\end{table}

We randomly selected 20\% of the images (80 images) as test data and used the rest of the images as training data (314 images). We augment the training dataset to generate sufficient data to train our proposed method. To make the the proposed method robust against, we construct a multi-scale pyramidal representation of each training image following the work \cite{boominathan2016crowdnet}. The multi-scale pyramidal representation of images includes scales of 0.4 to 1.3, incremented in steps of 0.1, times the original image resolution. Then, patches with size of $300\times300$ are cropped at random locations, which are followed by randomly flipping and adding Gaussian noise.

\subsection{Evaluation Metrics}

We use standard evaluation metrics, Mean absolute Error (MAE) and Root Mean Squared Error (RMSE), to measure the counting performance of the proposed method. Generally, MAE indicate the accuracy of the results and RMSE measures the robustness. These two metrics are defined as follows:

\begin{eqnarray}
MAE=\frac{1}{N}\sum_{i=1}^{N}|C_{i}^{pred}-C_{i}^{GT}|\label{eq:MAE_eq}
\end{eqnarray}

\begin{eqnarray}
RMSE=\sqrt{\frac{1}{N}\sum_{i=1}^{N}|C_{i}^{pred}-C_{i}^{GT}|^2}\label{eq:RMSE_eq}
\end{eqnarray}

\noindent where $N$ is the number of test images, $C_{i}^{pred}$ and $C_{i}^{GT}$ are the estimated count and the ground truth count corresponding to the $i$th test image.

\subsection{Training Hyperparameter}

DeepStand network is trained end-to-end from scratch. The network parameters are initialized with Xavier initialization \citep{glorot2010understanding}. Adam optimizer \citep{kingma2014adam} with learning rate of 3e-4 and a mini-batch size of 24 is used to minimize the loss function defined in Equation \eqref{eq:eq_loss}. The learning rate is gradually decayed to 25e-6 during the training process. The network is trained 80,000 iterations on 93,258 image patches generated following the data augmentation procedure in \ref{sec:data_aug}.

\subsection{Comparison with State-of-the-art}

To evaluate the efficiency of our proposed method, we compare our method with five state-of-the-art models. These five models were originally proposed for crowd counting problem, but they are also applicable for other object counting problems, which are as follows:

\textbf{CSRNet:} proposed by \cite{li2018csrnet}, uses a fully convolutional architecture which includes truncated VGG-16 network backbone as the front-end feature extractor and a set of dilated convolutional layers as back-end to estimate the density map.

\textbf{SaCNN:} proposed by \cite{zhang2018crowd}, employs a scale-adaptive CNN network to cope with scale and perspective change in images. The SaCNN uses a backbone similar to VGG-16 architecture for feature extraction and merges feature maps from different layers of network to make the model robust against scale variation. 

\textbf{MSCNN:} proposed by \cite{zeng2017multi}, extracts scale-relevant features using a multi-scale CNN network. The MSCNN network consists of multiple Inception-like \citep{szegedy2015going} modules for multi-scale feature extraction.

\textbf{CrowdNet:} proposed by \cite{boominathan2016crowdnet}, uses a multi-column network architecture. The network consists of a deep CNN and a shallow CNN to predict density map. The deep CNN has a VGG-like network architecture and the shallow CNN includes three convolutional layers.

\textbf{DeepCrowd:} proposed by \cite{wang2015deep}, is a regression-based method which directly learns a mapping from the image patches to the count. DeepCrowd network includes five convolutional layers which are followed by two fully connected layers.

\subsection{Results}

This section reports the evaluation results and compares our proposed method with other  state-of-the-art methods for the task of corn stand counting. After having trained all methods, we evaluated their performance on the hold-out test data which includes 80 images of corn stands from growing stages V1 to V6. Table \ref{tab:result1} illustrate the stand counting performances of the proposed and comparison methods on the test data with respect to MAE and RMSE evaluation metrics.

\begin{table}[H]
    \centering
    \begin{tabular}{|c| c| c|}
    \hline
         Method & MAE & RMSE \\
         \hline
         \hline
        CSRNet \citep{li2018csrnet}& 2.23& 2.84\\
        \hline
          SaCNN \citep{zhang2018crowd}& 4.38& 5.25\\
        \hline
          MSCNN \citep{zeng2017multi}& 2.26 & 2.65 \\
        \hline
          CrowdNet \citep{boominathan2016crowdnet}& 4.49& 5.53\\
        \hline
          DeepCrowd \citep{wang2015deep}& 2.61& 3.13\\
        \hline
          DeepStand (our proposed) & \textbf{1.73}& \textbf{2.46}\\
        \hline
    \end{tabular}
    \caption{The counting performances of the proposed and comparison methods on the test data. }
    \label{tab:result1}
\end{table}

Table \ref{tab:result1} illustrates that our proposed stand counting method outperforms the other methods to varying extents. The MSCNN has a comparable performance with CSRNet and both perform better than other methods except our proposed method. The main reason for the good performance of CSRNet is because of utilizing dilated convolutional layers to aggregate the multi-scale contextual information. The good performance of MSCNN can also be attributed to the use of multi-scale features which make it robust against scale variation. The DeepCrowd method as a regression-based method outperformed the SaCNN and the CrowdNet methods. The proposed method performed better than other methods for a couple of reasons: (1) our proposed method merges feature maps from multiple scales of the network to cope with scale variation, and (2) the use of deconvolution layers for up-sampling the feature maps increases the quality of the predicted density maps. 

Even though the counting performance of the CSRnet, MSCNN, DeepCrowd, and SaCNN are good, the performances of these methods are not satisfactory when the whole image is fed to these methods at once. As a result, an input image should be cropped into some non-overlapping patches and fed to these methods which increases their inference time. Our proposed method and CrowdNet take the whole image and predict the density map in a single forward path, which makes them considerably faster than other methods. 

\begin{figure}[H]
    \centering
    \includegraphics[scale=0.045]{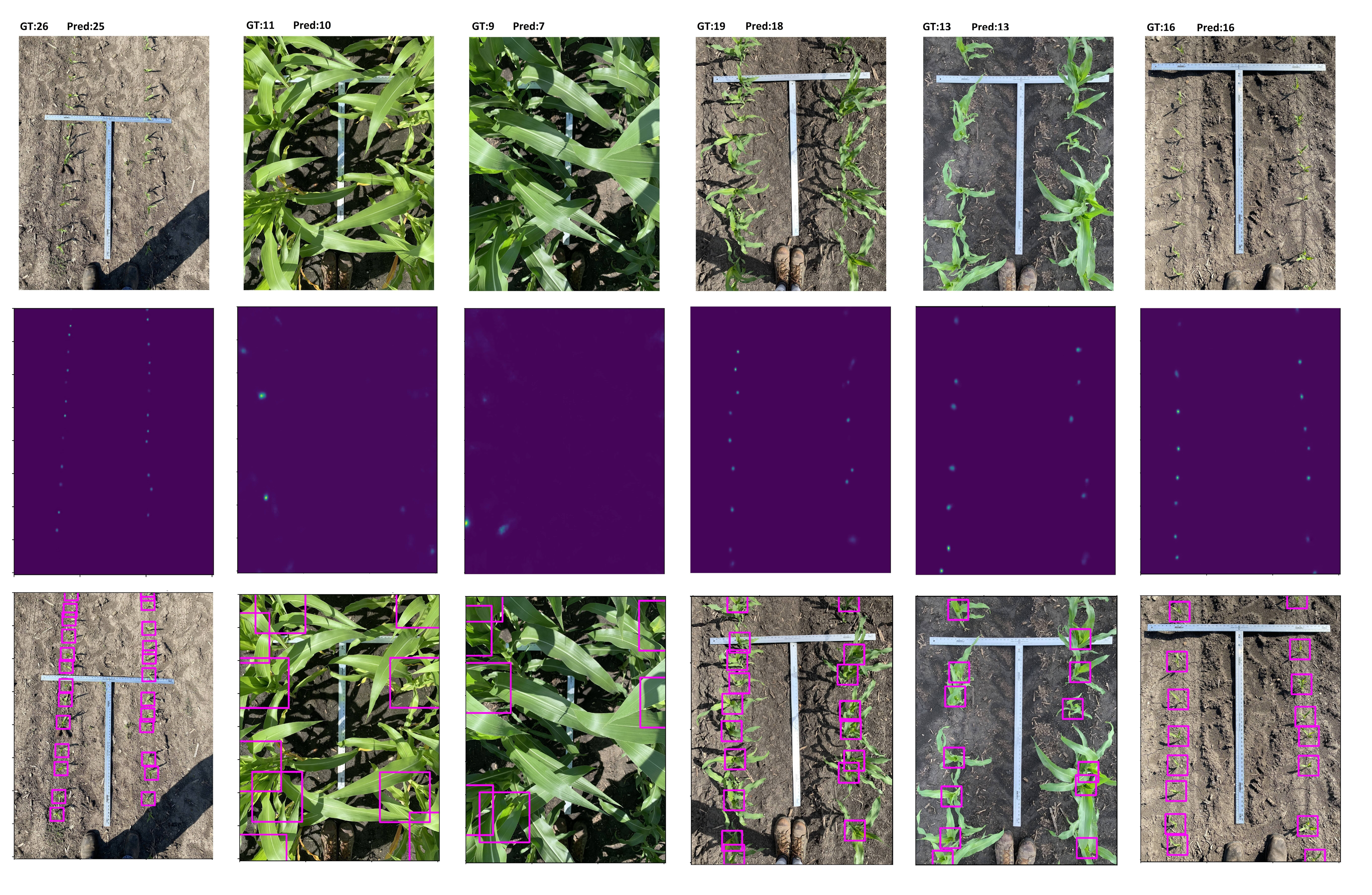}
    \caption{Visual results of our proposed stand counting method. The first, second and third rows indicate, respectively, input images, estimated density maps, and detected corn stand in images. The figure is best viewed in color.}
    \label{fig:result}
\end{figure}

Figure \ref{fig:result} visualizes some stand counting results of our proposed method including original image, predicted density map, and the detected corn stands in an image.

In order to see how the growing stages (VE to V6) affect the counting performance of the proposed method, we divide the test data based on the growing stages into three classes of VE-V1, V2-V4, and V5-V6 and report the counting performances of the proposed and comparison methods on these classes. As shown in Table \ref{tab:result_class}, the proposed method has a consistently low error across all growing stages which indicates the robustness of our proposed method. 
The results also indicate that the highest counting error of all methods except MSCNN belong to the stage V5-V6 due to the occlusion and background clutter.

\begin{table}[H]
    \centering
    \begin{tabular}{|c| c| c|c| c|c| c|}
    \hline
   \multirow{ 3}{*}{Method} &  \multicolumn{6}{|c|}{Growing stage}\\
     \cline{2-7}
         &  \multicolumn{2}{|c|}{VE-V1} & \multicolumn{2}{|c|}{V2-V4}&\multicolumn{2}{|c|}{V5-V6} \\
         \cline{2-7}
           & MAE & RMSE &  MAE & RMSE&  MAE & RMSE\\
         \hline
         
        CSRNet \citep{li2018csrnet}& 2.34& 3.12&1.51&\textbf{1.89}&3.04&3.56\\
        \hline
          SaCNN \citep{zhang2018crowd}& 5.87& 6.28 & 2.25&2.49&6.24&6.90\\
        \hline
          MSCNN \citep{zeng2017multi}& 2.72&2.98&2.37&2.76&\textbf{1.91}&\textbf{2.32} \\
        \hline
          CrowdNet \citep{boominathan2016crowdnet}&3.71&4.88 &4.43& 4.94 & 5.97&6.41\\
        \hline
          DeepCrowd \citep{wang2015deep}& 3.23&3.67&2.57&2.91&2.39&3.11\\
        \hline
          DeepStand (our proposed) & \textbf{1.0}& \textbf{1.31}& \textbf{1.39}& 1.94& 2.47& 3.30\\
        \hline
    \end{tabular}
    \caption{The counting performances of the proposed and comparison methods on the test data across different growing stages.  }
    \label{tab:result_class}
    
\end{table}

\section{Discussion}

In this paper, we presented a deep learning based approach named DeepStand for corn stand counting problem. The proposed method adopts a truncated VGG-16 network as a backbone feature extractor and merges multiple feature maps with different scales to make the network robust against scale variation. Finally, DeepStand uses a set of deconvolutional layers as a back-end to up-sample the network output. Our extensive experimental results indicate that our proposed method can successfully count and subsequently detect corn stands regardless of the image scale and the lighting condition. Our proposed method also outperformed other state-of-the-art methods commonly used in similar tasks.

\section*{Conflicts of Interest}

The authors declare no conflict of interest.



\bibliographystyle{elsarticle-harv}  
\bibliography{references}

\end{document}